\pdfoutput=1

\documentclass[11pt]{article}

\usepackage[table]{xcolor}
\definecolor{Gray}{gray}{0.9}
\usepackage{booktabs}
\usepackage{tabularx}
\usepackage{graphicx}
\usepackage{mathtools}
\usepackage{amsfonts}
\usepackage{acl}

\usepackage{times}
\usepackage{latexsym}

\usepackage{relsize}

\newcommand{\dashrule}[1][black]{%
  \color{#1}\rule[\dimexpr.5ex-.2pt]{4pt}{.4pt}\xleaders\hbox{\rule{4pt}{0pt}\rule[\dimexpr.5ex-.2pt]{4pt}{.4pt}}\hfill\kern0pt%
}

\usepackage[T1]{fontenc}

\usepackage[utf8]{inputenc}

\usepackage{microtype}

\title{IDIAPers @ Causal News Corpus 2022: Efficient Causal Relation Identification Through a Prompt-based Few-shot Approach
}

\author{Sergio Burdisso$^{*,1,2}$, Juan Zuluaga-Gomez$^{1,3}$, Esaú Villatoro-Tello$^{1,5}$, Martin Fajcik$^{1,4}$\\ 
\textbf{Muskaan Singh}$^{1}$\textbf{,} \textbf{Pavel Smrz}$^{4}$\textbf{,} \textbf{Petr Motlicek}$^{1}$\\
        $^{1}$Idiap Research Institute, Martigny, Switzerland\\
     $^{2}$Universidad Nacional de San Luis (UNSL), San Luis, Argentina\\
     $^{3}$Ecole Polytechnique Fédérale de Lausanne, Switzerland\\
     $^{4}$Brno University of Technology, Brno, Czech Republic\\
     $^{5}$Universidad Autónoma Metropolitana Unidad Cuajimalpa, Mexico City, Mexico\\
     $^{*}$\emph{corresponding author: sergio.burdisso@idiap.ch}
     }

\begin{document}
\maketitle

\begin{abstract}
In this paper, we describe our participation in the subtask 1 of CASE-2022, Event Causality Identification with Casual News Corpus.
We address the Causal Relation Identification (CRI) task by exploiting a set of simple yet complementary techniques for fine-tuning language models (LMs) on a small number of annotated examples (i.e., a \textit{few-shot} configuration).
We follow a prompt-based prediction approach for fine-tuning LMs in which the CRI task is treated as a masked language modeling problem (MLM). This approach allows LMs natively pre-trained on MLM problems to directly generate textual responses to CRI-specific prompts.
We compare the performance of this method against  ensemble techniques trained on the entire dataset.
Our best-performing submission was fine-tuned with only 256 instances per class, 15.7\% of the all available data, and yet obtained the second-best precision ($0.82$), third-best accuracy ($0.82$), and an F1-score ($0.85$) very close to what was reported by the winner team ($0.86$).\footnote{Code available at \url{https://github.com/idiap/cncsharedtask}.}
\end{abstract}

\section{Introduction}
\label{sect:intro}

Causal relation identification aims to predict whether or not there exists a cause-effect relation between a pair of events mentioned in a given text. For example, in the sentence \textit{``\underline{Protests} spread to 15 towns and resulted in the \underline{destruction of property}''}, the automatic causal identification system must be able to realize that there is cause-effect relation between the events \textit{``protest''} and \textit{``destruction''}.

Hence, understanding causal relations within a text is an essential aspect of natural language processing (NLP) and understanding (NLU)~\cite{ayyanar2019causal,LI2021207, Fiona2022}. Once the causal information is identified within a text, such knowledge becomes beneficial for many other downstream NLP tasks, e.g., Information Extraction, Question Answering, Text Summarization \cite{ayyanar2019causal, man-etal-2022-event}.  
However, due to the ambiguity and diversity in written documents, causality identification is not easy and remains a challenging problem.

The Event Causality Identification with Causal News Corpus (CASE-2022) shared task \cite{,tan-etal-2022-event} addresses this problem on a recently created corpus named the Causal News Corpus (CNC) \cite{tan-EtAl:2022:LREC}. Contrary to previous existing causality corpora, the CNC dataset, manually annotated by experts, incorporates a broader set of causal linguistic constructions, i.e., not only limited to 
explicit constructions, resulting in a more challenging dataset. 

In this paper, we describe our followed methodology for addressing the causal event classification shared task (subtask 1) during the CASE-2022 competition~\cite{tan-etal-2022-event}.\footnote{We refer the reader to our standalone publication \cite{idiap_subtaskB} to know our results for subtask 2.} 
Our primary method, based on a \textit{few-shot} configuration, follows a prompt-based approach for fine-tuning the language model (LM). The intuitive idea of this approach is to allow the LM to directly auto-complete natural language prompts. Following this technique, we leverage the LM's knowledge and let it decide the correct label of the input sequence. Additionally, we evaluate the performance of ensemble techniques trained using the entire dataset available. Our results demonstrate that our few-shot, prompt-based, fine-tuning approach can generalize well even when using as few as 256 samples per class 
for training,  outperforming ensemble techniques trained with the entire dataset, as well as most of other teams' submissions.

The rest of the paper is organized as follows. Section \ref{sect:rw} describes relevant related work, Section \ref{sub-sec:prompt} describes the components of our main method, namely the prompt-based approach. Section \ref{sect:exps} describes the experimental setup, i.e., datasets, additional baselines, experiments configuration and obtained results. Finally, Section \ref{sect:conclusions} depicts our main conclusions and future work directions.

\begin{figure*}[t]
    \centering
    \includegraphics[width=1\textwidth]{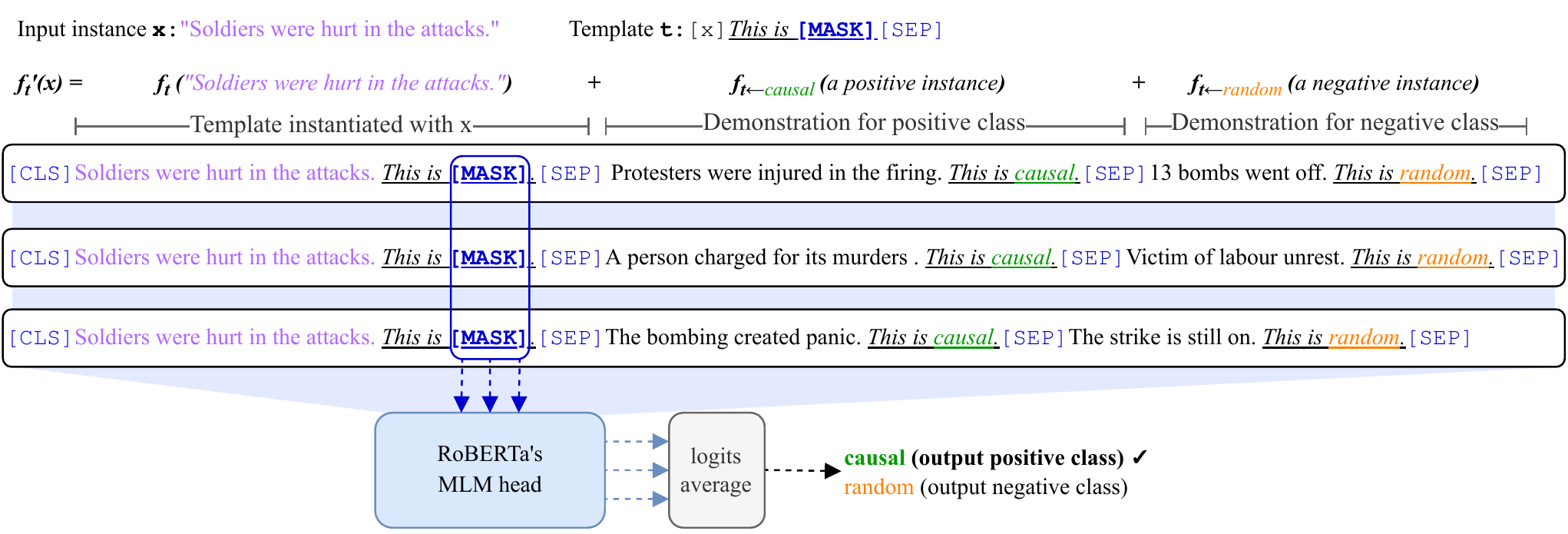}
    \caption{Augmented prompt-based classification for causality identification task.
    First, the input instance $x = \textit{``Soldiers were hurt in the attacks''}$ is converted into three different input prompts by applying $f_t'(x)$ three times. Then, these three prompts are given to a RoBERTa model, and one logit vector is obtained for each. These vectors are then averaged, and the word with the highest score, \emph{``causal''}, is selected. Finally, this word is mapped to its corresponding class, and $x$ is classified as positive. Note that, in this example, we have the following word-to-class label mapping $word(positive) = \textit{``causal''}$ and $word(negative) = \textit{``random''}$.}
    \label{fig:lm-bff}
\end{figure*}

\section{Related Work}
\label{sect:rw}


Previous work on causal relation identification varies from knowledge-based to deep neural network approaches (Deep-NN).  
Knowledge-based systems rely on linguistic patterns extracted using an exhaustive exploration of the data, where lexico-semantic and syntactic 
analysis lead to the identification of relevant structures and keywords that depict the presence of a causal relation in the text \cite{GarciaDaniela96, khoo-etal-2000-extracting}. Although interpretable, these methods 
require a lot of human effort to generate relevant patterns and result in models that are not readily applicable in different domains. 
 
Statistical machine learning (ML) approaches leave to the selected algorithm to find patterns in the data on the basis of the manual annotation. Traditionally, using different NLP tools, it is possible to compute various features for a given collection and apply any ML pipeline to train a causality relation classifier, e.g.,  \cite{rutherford-xue-2014-discovering, hidey-mckeown-2016-identifying}.
However, one main disadvantage of these techniques is the language dependency and error propagation of the NLP tools, e.g., syntactic parsers. 

Finally, recent approaches based on Deep-NN have become popular, given their powerful representation learning ability. Typical approaches include convolutional neural networks \cite{9028985}, long short-term memory networks \cite{LI2021207}, and pre-trained transformer-based LMs such as BERT~\cite{devlin-etal-2019-bert}, where following a standard fine-tuning approach makes possible the detection of causality relations \cite{Fiona2022, KhetanEtAl_2022, fajcik-etal-2020-fit}.
Normally, these methods involve high computational costs and large amounts of labeled data. However, in this work, we show that pre-trained LMs can still be effective even when fine-tuned with very few instances.

Contrary to previous work, we evaluate the effectiveness of very recent prompt-based prediction approaches under a \textit{few-shot} configuration for causal relation identification. 



\section{Prompt-Based Approach}
\label{sub-sec:prompt}

In the ``pre-train, prompt, and predict'' paradigm, unlike the standard ``pre-train and fine-tune'' paradigm, instead of adapting pre-trained LMs to downstream tasks via objective engineering,\footnote{\textit{Objective engineering} referes to both the pre-training and
fine-tuning stages of LMs~\cite{liu2021pre}.} downstream tasks are reformulated to look more like those solved during the LM pre-training phase~\cite{liu2021pre}.
More precisely, prompt-based prediction treats the downstream task as a masked language modeling problem, where the model directly generates a textual response (referred to as a \emph{label word}) to a given prompt defined by a task-specific \emph{template}~\cite{gao2020making}.
For instance, when identifying the sentiment of a movie review like ``I love this movie.'' we may continue with ``Overall, it was a \texttt{[MASK]} movie.'' and ask the LM to fill the mask with a sentiment-bearing word.
In this example, the original input text $x$ (``I love this movie.'') is modified using the \emph{template} ``\texttt{[x]} Overall, it was a \texttt{[MASK]} movie.'' into a textual string prompt $x'$ in which the mask will be filled with a \emph{label word}.
Some examples of \emph{label words} for this example could be ``fantastic'' or ``boring''.

In the case of classification tasks, in addition to defining a set of possible \emph{label words}, it is necessary to define a mapping between each one and the actual output labels. For instance, if labels $+$ and $-$ refer to positive and negative sentiment, respectively, ``fantastic'' in previous example could be mapped to output label $+$, and ``boring'' to $-$.

Formally, let $\mathcal{L}$ be a pre-trained language model, $f_t(x)$ a function that converts the input $x$ into a prompt by instantiating template $t$ which contains one \texttt{[MASK]} token, $mask$. Let $word: \mathcal{Y} \rightarrow \mathcal{W}$ be a mapping from the task label space, $\mathcal{Y}$, to the \emph{label words} set, $\mathcal{W}$.
Then, the classification task is converted to a \emph{masked language modeling} (MLM) task in which the probability of predicting class $y \in \mathcal{Y}$ is modeled as:
\begin{equation}
\label{eq:prompt}
    \begin{split}
    p(y|x) & = p(mask = word(y) | f_t(x)) =\\
    & = \frac{exp(\mathbf{w}_{word(y)} \cdot \mathbf{h}_{mask})}{\sum_{y'\in \mathcal{Y}}exp(\mathbf{w}_{word(y')} \cdot \mathbf{h}_{mask})},
    \end{split}
\end{equation}
where $\mathbf{h}_{mask}$ is the hidden vector of \texttt{[MASK]} and $\mathbf{w}_v$ denotes the vector encoding word $v$.
Note that when fine-tuning $\mathcal{L}$ to minimize the cross-entropy loss, the pre-trained weights $\mathbf{w}_v$ are re-used, and there's no need to introduce any new parameter. On the contrary, with standard fine-tuning a task-specific head, $softmax(\mathbf{W}_o \mathbf{h}_{\texttt{[CLS]}})$, has to be added, with new task-specific learnable parameters $\mathbf{W}_o \in \mathbb{R}^{|\mathcal{Y}|\times d}$, which increases the gap between pre-training and fine-tuning.

Hereafter 
we will refer to the "causal" and "non-causal" classes as ``positive'' ($+$) and ``negative'' ($-$) respectively. 
In addition, 
and following previous work by~\citet{gao2020making}, we append one answered prompt for each class to the input prompt as \emph{demonstrations}.\footnote{These~\emph {demonstrations}~\cite{gao2020making} are used to demonstrate the LM, in-context, how it should provide the answer to the input prompt.} 
More precisely, let $\mathcal{Y}=\{+,-\}$ be the set of labels for the binary causality identification task, let $t\leftarrow v$ be the template $t$ in which its \texttt{[MASK]} token has been filled with word $v$, and $w^y = word(y)$ the \emph{word label} for class $y \in\mathcal{Y}$, then we redefine $f_t(x)$ in \autoref{eq:prompt} as $f_t'(x)$ defined as:
\begin{equation}
\label{eq:demo}
    f_t'(x) = f_t(x)\mathbin\Vert f_{t\leftarrow w^+}(x^+)\mathbin\Vert f_{t\leftarrow w^-}(x^-)
\end{equation}
where $\mathbin\Vert$ is the string concatenation operator, and $x^y$ is an instance of class $y$ randomly sampled from the training set. 
\autoref{fig:lm-bff}, depicts an example of three different input prompts are shown by applying $f_t'(x)$ three times to the input instance $x$.

\noindent\textbf{Classification process:} the process is illustrated 
in \autoref{fig:lm-bff}. First, the input instance $x$ is converted into $d$ different input prompts by applying $f_t'(x)$, $d$ times. Then, each input prompt is given to the LM to obtain $d$ logit vectors holding the word scores for the mask in each prompt. A simple ensemble scheme is then applied by averaging all $d$ logit vectors, and the \emph{word label} with the highest score is selected, which is finally mapped to its corresponding class $y$ using mapping $word(y)$.


\begin{table*}[t!]
    \centering
    \begin{tabular}{l ccc ccc ccc cccc}
        \toprule
         &  & \multicolumn{2}{c}{\cellcolor{Gray} \textbf{Precision}} &  & \multicolumn{2}{c}{\cellcolor{Gray} \textbf{Recall}} &  & \multicolumn{2}{c}{\cellcolor{Gray} \textbf{Accuracy}} &  & \multicolumn{2}{c}{\cellcolor{Gray} \textbf{F1-Score}} \\
         \cline{3-4} \cline{6-7} \cline{9-10} \cline{12-13}
        Submission &  & \textit{dev} & \textit{test} &  & \textit{dev} & \textit{test} &  & \textit{dev} & \textit{test} &  & \textit{dev} & \textit{test} \\
        \midrule
        Ensemble-10m &  & \textbf{88.46} & 82.78 &  & 90.45 & 84.66 &  & \textbf{88.26} & 81.35 &  & \textbf{89.44} & 83.70 \\
        Prompt-256 &  & 85.49 & \textbf{82.80} &  & \textbf{92.70} & 87.50 &  & 87.30 & \textbf{82.64} &  & 88.95 & \textbf{85.08} \\
        Prompt-356e &  & 82.72 & 80.41 &  & 88.76 & \textbf{88.64} &  & 83.60 & 81.35 &  & 85.63 & 84.32 \\
        Prompt-1000 &  & 84.56 & 81.08 &  & 91.57 & 85.22 &  & 86.07 & 80.39 &  & 87.87 & 83.10 \\
        Ensemble-8p &  & 86.10 & 81.15 &  & 90.44 & 88.07 &  & 86.69 & 81.67 &  & 88.22 & 84.47 \\
        \bottomrule
    \end{tabular}
    \caption{Official performance metrics in percentages (\%) from the selected methods in dev and test partitions of the Causal News Corpus.}
    \label{tab:exp_submissions}
\end{table*}

\noindent\textbf{Training and model selection}: for developing our prompt-based models, we performed a simplified version of the process described in previous work by \citet{gao2020making}.
Namely, we carried out the following six steps:

\textbf{Step 1}: we created a new training set, $\tau_k$, by extracting $k$ instances per class from the original train partition, and used the remaining $2925 - 2\times k$ instances as a large evaluation set $\delta_{T-k}$ (dataset stats are given in \autoref{tab:dataset}).


\textbf{Step 2}: in order to add \emph{demonstrations} to a given input $x$ (see \autoref{eq:prompt}), we uniformly sampled $x^-$ and $x^+$ from the top-$50\%$ most similar instances in $\tau_k$.\footnote{We tested different percentages, however $50\%$ was the best-performing one.} To do so, we pre-computed the sentence embeddings of training instances using a pre-trained SBERT \cite{reimers2019sentence} model, and cosine distance was used as a similarity metric.

\textbf{Step 3}: using \textit{``causal''} and \textit{``random''} as \emph{word labels},\footnote{We performed some simple preliminary tests using different words like ``coincidence'', ``choice'', ``causal'', ``cause'', with few trivial hand-crafted templates (e.g. $\text{``\texttt{[x]} \textit{It was} \texttt{[MASK]}''}$), from which ``random'' and ``casual'' where selected.} the next step was to generate candidate templates automatically using T5. First, each training instance $x$ of class $y$ in $\tau_{k}$ was converted to  ``$\texttt{[x]}\texttt{<P>}word(y)\texttt{<S>}$'' where $\texttt{<P>}$ and $\texttt{<S>}$ are T5 mask tokens, and used a 100 wide beam search to decode multiple template candidates by filling $\texttt{<P>}$ and $\texttt{<S>}$ tokens.

\textbf{Step 4}: next step was sorting all 100 final candidate templates by F1 score. However, since this is a time-consuming step, a subset of the evaluation set was used by sampling $256$ unique positive and negative instances from $\delta_{T-k}$. Note that no fine-tuning is used at this point, just the out-of-the-box pre-trained LM.

\textbf{Step 5}: we selected the top-10 best-performing templates as final candidates. For each candidate template we fine-tuned the LM as a MLM task (see \autoref{eq:prompt}) on the training set, $\tau_{k}$, evaluating it on the complete evaluation set, $\delta_{T-k}$.

\textbf{Step 6}: finally, the model with the best F1 score on the official dev set was selected as a candidate for submission ---we also checked that the F1 score on $\delta_{T-k}$ was among the first ones too (if not first).
Note that in this step we're evaluating the model on unseen data since the official dev set is being used as an unofficial test set.

\

The above process was repeated varying the number $k$ of training instances, with $k=256$, $356$, $512$, and $1000$;\footnote{Inspired by evidence showing a performance saturation when $k=256$ (Figure 3 in \citet{gao2020making}), compared to standard fine-tuning on the entire dataset, we decided to start from this value.} the number $d$ of input prompts to ensemble during classification stage, with $d$ from $1$ to $9$; and using  RoBERTa (large and base), and DeBERTa V3 (base) as pre-trained LMs.
In step 5, models were fine-tuned for a maximum of $1000$ steps using AdamW~\cite{loshchilov2018decoupled} optimizer ($\beta_1{=}0.9, \beta_2{=}0.999, \epsilon{=}1\mathrm{e}{-8}$) with a learning rate of $\gamma{=}1\mathrm{e}{-5}$ with no weight decay ($\lambda{=}0$). Models were evaluated every 100 steps and check-pointed when new best F1 scores were obtained.



\section{Results \& Discussion}
\label{sect:exps}
In this section we provide the details of the employed dataset, a set of additional experiments based on recent ensemble techniques, and the final configuration of our submitted runs to the subtask 1 of CASE 2022. 

\subsection{Dataset}
\label{subsect:dataset}
As mentioned earlier, the main goal of subtask 1 of CASE-2022 is to classify whether or not a given sentence contains a \textit{cause-effect} relation. Thus, systems have to be able to predict \textit{Causal} or \textit{Non-causal} labels per sentence. Table \ref{tab:dataset} contains a few statistics regarding the distribution of the classes in the \textit{train, dev}, and \textit{test} partitions. 

\begin{table}[]
    \centering
    \begin{tabular}{lcccc}
    \toprule
    Label & {\cellcolor{Gray}\textbf{Train}} & {\cellcolor{Gray}\textbf{Dev}} & {\cellcolor{Gray}\textbf{Test}} & {\cellcolor{Gray}\textbf{Total}}\\
    \midrule
    \textit{Causal} & 1603 & 178 & 176 & 1957 \\
    \textit{Non-causal} &  1322 & 145 & 135 & 1602\\
    \cline{2-5}
    Total: & 2925 & 323 & 311 & \textbf{3559}\\
    \bottomrule
    \end{tabular}
    \caption{Number of positive (causal) and negative (non-causal) instances in the \textit{train, dev,} and \textit{test} sets of the shared task. We refer the interested reader to \cite{tan-etal-2022-event} to know more details about the data and the labeling process.}
    \label{tab:dataset}
\end{table}

\subsection{Ensemble-based Approach}
\label{subsect:ensemble}
We also performed several ensembles of different fine-tuned LMs to increase the generalization and compensate for the overfitting of the models.
We followed the approach described in~\citet{fajcik2019but}, called \textit{TOP-N} fusion. In this formulation, we first define a \textit{set} of $M$ pre-trained LMs, varying the training seed. \textit{TOP-N} fusion starts by choosing one uniformly random model from the \textit{set}, which is added to the ensemble. Next, it randomly shuffles the rest of the models and tries adding them into the ensemble once, as long as the F1 score improves. Each time a model is added to the ensemble, its performance gets measured. The model would stay in the ensemble only and only if it improved the overall performance. This aims at an iterative optimization of the ensemble's F1 score by averaging the output probabilities. As the selection process is stochastic, we repeat the process $N{=}10000$ times. We construct a new ensemble for each iteration, independently of the previous ones. Finally, we select the best performing ensemble for submission. Further details are given in \autoref{appendix:ensemble} (\autoref{fig:ensemble}). 


\subsection{Official Submissions}
\label{sub-sec:submissions}

Next, we describe each one of our submissions:

\noindent \textbf{Ensemble-10m:} ensemble model described in \autoref{subsect:ensemble} with 10 final models obtained from a set of 150 initial ones (50 fine-tuned \textit{bert-base-cased}, \textit{roberta-base}, and \textit{deberta-v3-base} models).

\noindent \textbf{Prompt-256:} prompt-based \textit{roberta-large} model with $k{=}256$ training instances per class, $d{=}3$ input prompts to ensemble during classification stage; and template $t = \text{``\texttt{[x]} \textit{This is not} \texttt{[MASK]}''}$.

\noindent \textbf{Prompt-1000:} The same previous model but with $t = \text{``\texttt{[x]} \textit{There were no} \texttt{[MASK]}\textit{ities in this}''}$, $k{=}1000$, and $d{=}1$.

\noindent \textbf{Ensemble-8p:} ensemble model described in \autoref{subsect:ensemble} with 8 final models obtained from the top-50 best performing prompt-based models as the initial set.

\noindent \textbf{Prompt-356e:} three prompt-base models trained with $k{=}356$ instances. The first two models have the same template as \emph{Prompt-1000} but with $d{=}2$ and $3$, respectively. The third one uses the  template $t = \text{``\texttt{[x]} \textit{The incident is not} \texttt{[MASK]}''}$ with $d{=}1$.\footnote{Note that these prompts, as well as previous ones, were automatically generated as described in \autoref{sub-sec:prompt}.}
Finally, a simple majority voting ensemble among these three models generates the output. 


\subsection{Results}
\label{sub-sec:results}
\autoref{tab:exp_submissions} shows the official results, both in dev and test partitions, for our five submissions. As expected, the ensemble of several LMs (Ensemble-10m) was able to obtain outstanding performance across several metrics during the validation phase (i.e., dev partition\footnote{
We further performed a 5-cvf experiment on six different architectures, see the results on \autoref{tab:exp_k-fold} in \autoref{appendix:baseline_results}.}).
However, the performance dropped significantly in the test partition (F1$=89.44$ $\rightarrow$ F1$=83.70$).
On the contrary, our prompt-based approach trained on 256 instances per class (Prompt-256) could generalize better on the test partition. Such submission obtained 2nd place in terms of precision (82.80\%), 3rd in accuracy (82.64\%), and 5th in F1 (85.08\%) ---the best F1 was $86.19\%$.
However, the main advantage of our approach is that it allows the LM to be trained in a few-shot setting, making it harder for the model to overfit the data.
Moreover, most of the available data can be kept and used for measuring the generalization power of the model instead.
For instance, our best-performing model (Prompt-256) was fine-tuned only on $15.7\%$ of all available data,\footnote{i.e. train + dev sets in \autoref{tab:dataset}} allowing the remaining $84.3\%$ to be used for evaluation and model selection ($74.3\%$ as evaluation set and $10\%$ as our own test set).
Therefore, model selection choice is more robust since the risk of performance drop on unseen data, such as the official test set, is expected to be lower.

\section{Conclusions}
\label{sect:conclusions}

This paper describes our participation in the CASE-2022 subtask 1. 
Our proposed approach uses a few-shot configuration in which a prompt-based model is fine-tuned using \emph{only 256 instances} per class and yet was able to obtain remarkable results among all 16 participant teams. The comparison against traditional fine-tuning techniques, ensemble approaches, as well as the other participating models, show the potential of the proposed approach for better generalizing the posed task.

For future work, we plan to perform further ablation studies when we have access to test set ground truth labels. For instance, measuring the dev-to-test performance drop in relation to $k$ or the robustness against different training and demonstration sampling given a fixed $k$.
\section*{Acknowledgements}
This work was supported by CRiTERIA, an EU project, funded under the Horizon 2020 programme, grant agreement no. 101021866 and the Ministry of Education, Youth and Sports of the Czech Republic through the e-INFRA CZ (ID:90140).
Esaú Villatoro-Tello, was supported partially by Idiap, SNI CONACyT, and UAM-Cuajimalpa Mexico.

\bibliography{references}

\bibliographystyle{acl_natbib}

\appendix
\newpage
\section{Baseline results}
\label{appendix:baseline_results}

We performed standard cross-entropy fine-tuning on six different pre-trained LMs (see first column in~\autoref{tab:exp_k-fold}) to produce baselines. We perform 5-fold cross-validation for each architecture following the partitions proposed in~\citet{tan-EtAl:2022:LREC}. Each system is fine-tuned on the sequence classification task to discriminate between casual and non-causal text input sequences. We report the mean and standard deviation (mean $\pm$ std) on the official development set over several metrics, see~\autoref{tab:exp_k-fold}. 

During experimentation, we use the same learning rate of $\gamma=5\mathrm{e}{-5}$ with a linear learning rate scheduler. Dropout is set to $dp=0.1$ for the attention and hidden layers, while Gaussian Error Linear Units (GELU) is used as activation function~\cite{hendrycks2016gaussian}. We fine-tune each model with an effective batch size of 32 for 50 epochs with AdamW \cite{loshchilov2018decoupled} optimizer ($\beta_1{=}0.9, \beta_2{=}0.999$, $\epsilon{=}1\mathrm{e}{-8}$). We noted that \textit{deberta-v3-base} performed systematically better in all metrics as shown in~\autoref{tab:exp_k-fold}.

\begin{figure*}[t]
    \centering
    \includegraphics[width=1\textwidth]{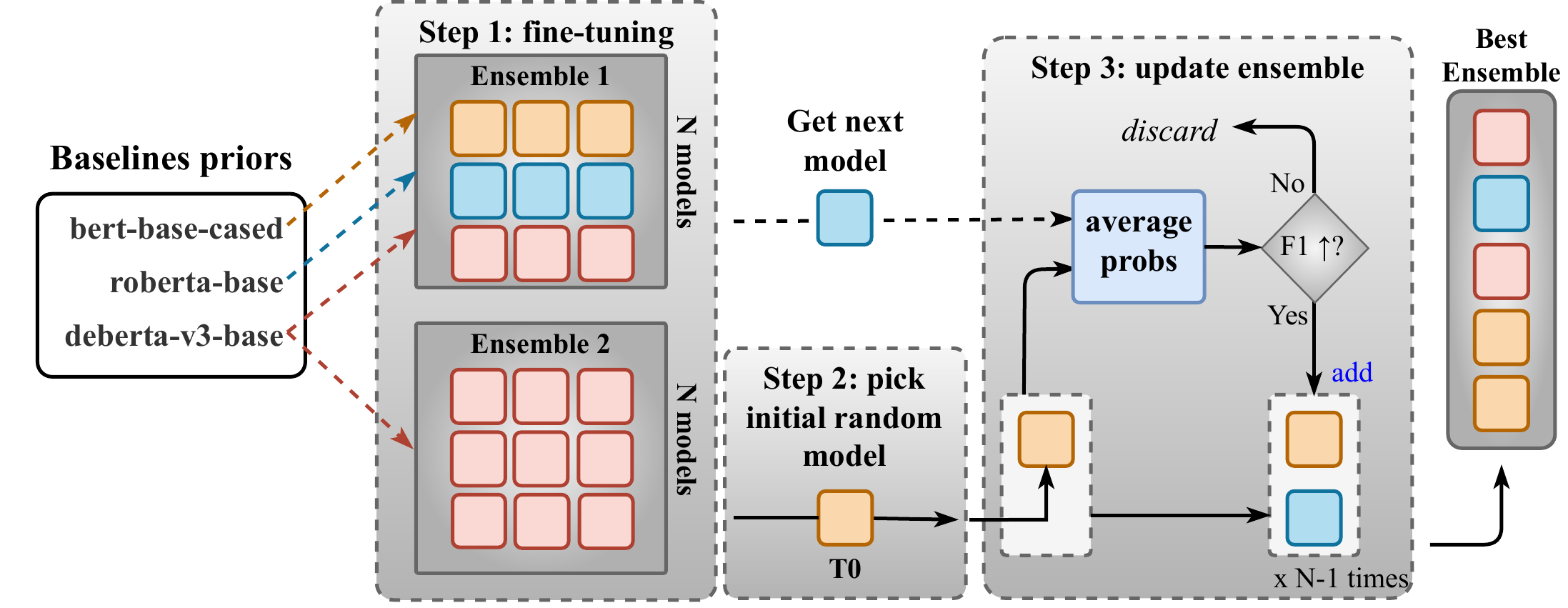}
    \caption{Our proposed method to ensemble $N$ fine-tuned LMs, based on~\citet{fajcik2019but} approach. We fine-tune several LMs by modifying only the training seed. Our implementation uses the sequence classification task from HuggingFace toolkit \cite{wolf2020transformers,lhoest2021datasets}.}
    \label{fig:ensemble}
\end{figure*}

\begin{table*}[t]
    \centering
    \scalebox{0.9}{
    \begin{tabular}{l c c c c c} 
        \toprule
        \textbf{Model} & \multicolumn{1}{c}{\cellcolor{Gray} \textbf{Precision}} & \multicolumn{1}{c}{\cellcolor{Gray} \textbf{Recall}} & \multicolumn{1}{c}{\cellcolor{Gray} \textbf{Accuracy}} & \multicolumn{1}{c}{\cellcolor{Gray} \textbf{F1-score}} & \multicolumn{1}{c}{\cellcolor{Gray} \textbf{Reference}} \\
        \midrule
        
        \texttt{bert-base-cased} & 83.52 $\pm$ 1.01 & 87.88 $\pm$ 3.08 & 79.68 $\pm$ 1.83 & 81.03 $\pm$ 1.20 &  {\small\cite{devlin-etal-2019-bert}} \\
        \texttt{bart-base} & 84.21 $\pm$ 0.88 & 87.80 $\pm$ 2.26 & 80.99 $\pm$ 2.19 & 81.98 $\pm$ 0.95 &  {\small\cite{lewis2020bart}} \\
        \texttt{roberta-base} & 85.13 $\pm$ 1.11 & 87.86 $\pm$ 2.41 & 82.66 $\pm$ 2.35 & 83.21 $\pm$ 1.10 &  {\small\cite{liu2019roberta}}\\
        \texttt{distilroberta-base} & 84.41 $\pm$ 1.20 & 88.05 $\pm$ 1.69 & 81.12 $\pm$ 2.09 & 82.22 $\pm$ 1.12 & {\small\cite{sanh2019distilbert}} \\
        \texttt{deberta-base} & 82.67 $\pm$ 2.76 & 85.74 $\pm$ 2.72 & 80.32 $\pm$ 6.49 & 80.31 $\pm$ 3.44 &  {\small\cite{he2021deberta}}\\
        \texttt{deberta-v3-base} & \textbf{85.87 $\pm$ 1.18} & \textbf{88.88 $\pm$ 1.74} & \textbf{83.18 $\pm$ 3.16} & \textbf{84.00 $\pm$ 1.18} &  {\small\cite{he2021debertav3}}\\
        
        \bottomrule
    \end{tabular}
    }
    \caption{Mean and standard deviation (mean $\pm$ std) of different metrics on the dev set using a 5-fold cross validation scheme on the CNC dataset. We report results for six different architectures of pre-trained LMs.}
    \label{tab:exp_k-fold}
\end{table*}

\section{Ensembling}
\label{appendix:ensemble}

We compose ensembles before submission to leaderboard in two manners. \texttt{Ensembling-type-1} and \texttt{Ensembling-type-2}: 

\begin{itemize}
    \item \texttt{Ensembling-type-1}: we define a \textit{set} of models, which contains only baseline LMs fine-tuned on the sequence classification task (see~\autoref{tab:exp_k-fold}). We fine-tune 50 LMs for each architecture from first column of~\autoref{tab:exp_k-fold}. Next, we run our \textit{TOP-N} fusion algorithm (see~\autoref{subsect:ensemble}) with the \textit{set} of models previously defined. The model submitted with \texttt{Ensembling-type-1} is \mbox{\textbf{Ensemble-10m}}, reporting its performance in~\autoref{tab:exp_submissions}.
    \item \texttt{Ensembling-type-2}, we define a \textit{set} of models containing prompt-based LMs. We select the top models for leaderboard submission. The overall process for ensembling is illustrated in~\autoref{fig:ensemble}. Even though the figure only depicts our first approach (explained above), we perform exactly the same with the prompt-based models explained in~\autoref{sub-sec:prompt}. The model submitted with \mbox{\texttt{Ensembling-type-2}} is \textbf{Ensemble-8p}, reporting its performance in~\autoref{tab:exp_submissions}.
\end{itemize}

\begin{table}[ht]
    \centering
    \begin{tabular}{ll}
        \toprule
        Model & F1-score (\%) \\
        \midrule
        \texttt{bert-base-cased} & 85.15 \\
        \texttt{roberta-base} & 86.76 \\
        \texttt{deberta-v3-base} & 89.69 \\
        \midrule
        \textit{\textbf{Ensemble-10m}}$^{\dagger}$  & \textbf{89.7} \\
        \bottomrule
    \end{tabular}
    \caption{Obtained F1-scores on the \textit{dev} partition of subtask 1 of the Causal News Corpus. Results depict the top performance of three models that belong to the \textit{\textbf{Ensemble-10m}} configuration. The last row corresponds to an ensemble model composed of ten independent LMs, namely, six \textit{deberta-v3-base}, two \textit{bert-base-cased}, and two \textit{roberta-base}. More details about the ensemble construction are described in ~\autoref{subsect:ensemble} and~\citet{fajcik2019but}.}
    
    \label{tab:exp_ensemble_only_lms}
\end{table}

\textbf{Details about the ensemble:} we select the best ensembles based on its F1-score performance on the dev set. For example, in~\autoref{tab:exp_ensemble_only_lms} we list the performance of the \texttt{Ensembling-type-1} system (i.e., \mbox{\textbf{Ensemble-10m}}) we used for our submission in the leaderboard.





\end{document}